\documentclass[runningheads]{llncs}
\usepackage{graphicx}
\usepackage{tikz}
\usepackage{comment}
\usepackage{amsmath,amssymb} % define this before the line numbering.
\usepackage{booktabs}
\usepackage{multirow}
\usepackage{multicol}
\usepackage{color}
\usepackage{xspace}
\usepackage{xcolor}
\usepackage{subcaption}
\usepackage[ruled,vlined]{algorithm2e}
\usepackage{url}
\usepackage{hyperref}

\usepackage[width=122mm,left=12mm,paperwidth=146mm,height=193mm,top=12mm,paperheight=217mm]{geometry}

\newcommand{\lblfig}[1]{\label{fig:#1}}
\newcommand{\ours}{LDET\xspace}

\makeatletter
\DeclareRobustCommand\onedot{\futurelet\@let@token\@onedot}
\def\@onedot{\ifx\@let@token.\else.\null\fi\xspace}
\def\eg{\emph{e.g}\onedot} 
\def\ie{\emph{i.e}\onedot} 
 
\def\etc{\emph{etc}\onedot} 
\def\wrt{w.r.t\onedot} 
 
\def\etal{\emph{et al}\onedot}

\begin{document}

%\linenumbers

\pagestyle{headings}
\mainmatter

\title{Learning to Detect Every Thing \\in an Open World} % Replace with your title

%\begin{comment}
%\titlerunning{ECCV-22 submission ID \ECCVSubNumber} 
%\authorrunning{ECCV-22 submission ID \ECCVSubNumber} 
\author{%
  Kuniaki Saito$^{1}$ \ Ping Hu$^{1}$ \ Trevor Darrell$^{2}$ \ Kate Saenko$^{1,3}$\\}
\institute{$^{1}$Boston University \ $^{2}$University of California, Berkeley \ $^{3}$MIT-IBM Watson AI Lab\vspace{.7em}\\}
\authorrunning{Saito et al.}

%\end{comment}
%******************

\begin{comment}
\titlerunning{Abbreviated paper title}
\author{First Author\inst{1}\orcidID{0000-1111-2222-3333} \and
Second Author\inst{2,3}\orcidID{1111-2222-3333-4444} \and
Third Author\inst{3}\orcidID{2222--3333-4444-5555}}
%
\authorrunning{F. Author et al.}
\institute{Princeton University, Princeton NJ 08544, USA \and
Springer Heidelberg, Tiergartenstr. 17, 69121 Heidelberg, Germany
\email{lncs@springer.com}\\
\url{http://www.springer.com/gp/computer-science/lncs} \and
ABC Institute, Rupert-Karls-University Heidelberg, Heidelberg, Germany\\
\email{\{abc,lncs\}@uni-heidelberg.de}}
\end{comment}
%******************
\maketitle

\begin{figure}
    \centering
    \vspace{-10mm} %10
   \includegraphics[width=\linewidth]{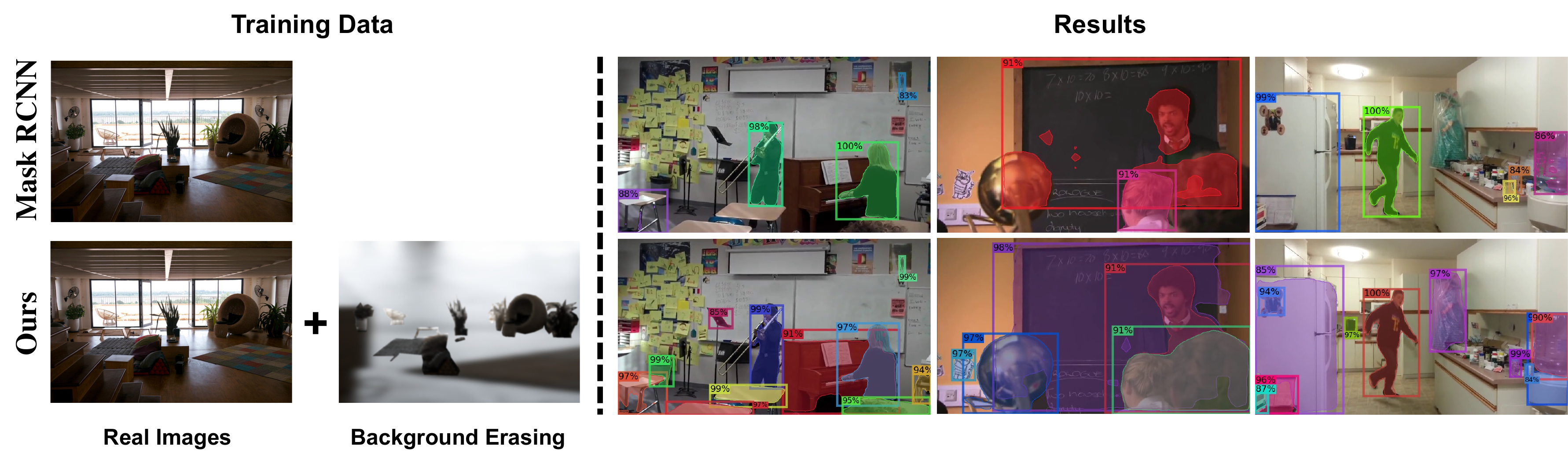}
    \vspace{-7mm} %7
        \caption{\small In an open world detection task, the model must locate and segment all objects in the image irrespective of categories used for training. \textbf{Left}: We propose a new multi-domain training scheme using real images and augmented images with ``erased'' background. \textbf{Right}: When training on COCO~\cite{coco} and testing on UVO~\cite{wang2021unidentified}, our detector correctly localizes many objects that are not labeled in COCO with the help of our new data augmentation and training scheme.}
   % \caption{\textbf{Mask RCNN (top row) detects fewer objects than our approach (bottom row) in open-world instance segmentation.} In this task, the model must locate and segment all objects in the image irrespective of categories used for training. Here both detectors are trained on COCO~\cite{coco} and tested on UVO~\cite{wang2021unidentified}.
   % Our detector correctly localizes many objects that are not labeled in COCO with the help of a new data augmentation method and training scheme.}
    \lblfig{teaser}
    \label{fig:examples_result_uvo}
    \vspace{-10mm}
\end{figure}

\begin{abstract}
 Many open-world applications require the detection of novel objects, yet state-of-the-art object detection and instance segmentation networks do not excel at this task. The key issue lies in their assumption that regions without any annotations should be suppressed as negatives, which teaches the model to treat any unannotated (hidden) objects as background. To address this issue, we propose a simple yet surprisingly powerful data augmentation and training scheme we call Learning to Detect Every Thing (LDET). To avoid suppressing hidden objects, we develop a new data augmentation method, BackErase, which pastes annotated objects on a background image sampled from a small region of the original image. Since training solely on such synthetically-augmented images suffers from domain shift, we propose a multi-domain training strategy that allows the model to generalize to real images. 
 \ours leads to significant improvements on many datasets in the open-world instance segmentation task, outperforming baselines on cross-category generalization on COCO, as well as cross-dataset evaluation on UVO, Objects365, and Cityscapes.
 \vspace{-3mm}
\end{abstract}

\vspace{-5mm}
\section{Introduction}
\label{sec:intro} 
\begin{figure}[t]
    \centering
    \includegraphics[width=0.7\linewidth]{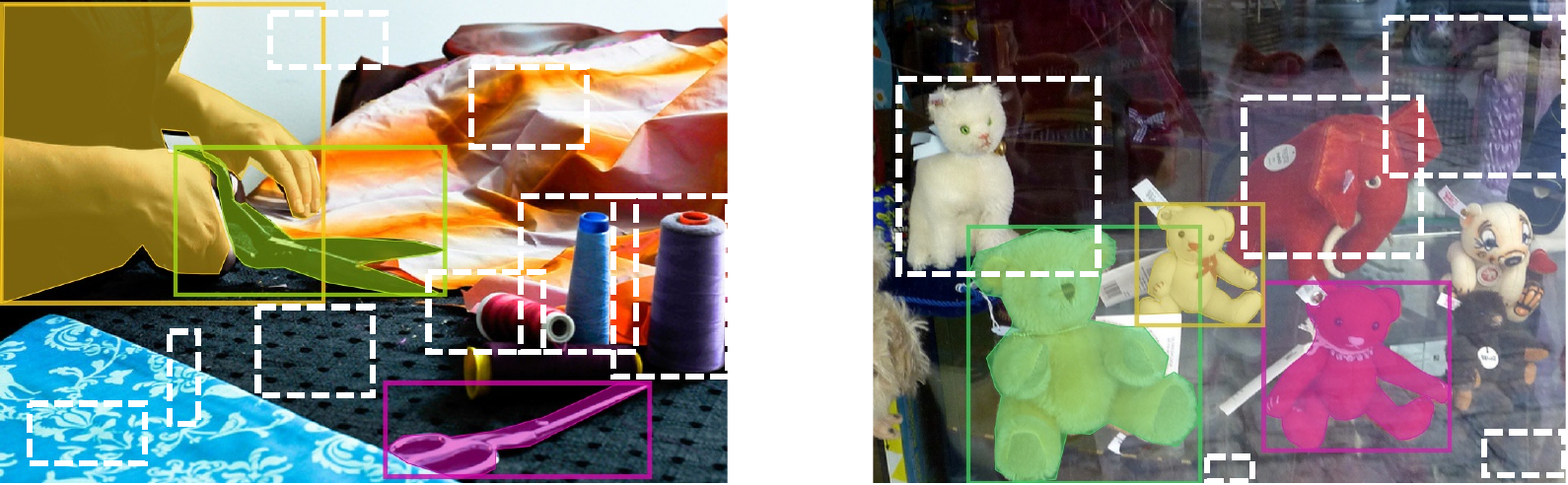}
    \vspace{-2mm}
    \caption{\small{\textbf{Hidden objects:} Existing datasets do not exhaustively label all objects, leading to detectors that are ill-prepared to propose boxes for long-tail categories. Colored boxes are annotated boxes while white-dashed boxes are potential background regions.
    Many white-dashed regions locate objects, but are regarded as background during training. This can suppress the objectness of novel objects.}}
    \label{fig:motivation}\vspace{-3mm}
\end{figure}

\vspace{-3mm}
Humans routinely encounter new tools, foods, or animals, having no problem perceiving the novel objects as \textit{objects} despite having never seen them before. 
Unlike humans, current state-of-the-art detection and segmentation methods ~\cite{maskrcnn,ssd,yolo,faster,lin2017focal} have difficulty recognizing novel objects as \textit{objects} because these methods are designed with a closed-world assumption. Their training aims to localize known (annotated) objects while regarding unknown (unannotated) objects as \textit{background}. This causes the models to fail in locating novel objects and learning general \textit{objectness}. 
One way to deal with this challenge is to create a dataset with an exhaustive annotation of every single object in each image. However, creating such datasets is very expensive. In fact, many public datasets~\cite{coco,everingham2010pascal,zhou2017scene} for object detection and instance segmentation do not label all objects in an image (Fig.~\ref{fig:motivation}). 

Failing to learn general objectness can cause issues in many applications. For instance, embodied AI (e.g., robotics, autonomous driving) requires localizing objects unseen during training. 
Autonomous driving systems need to detect novel objects in front of the vehicle to avoid accidents though identifying the category is not necessarily required. In addition, zero-shot, and few-shot detection have to localize objects unseen during training.
Open-world instance segmentation~\cite{wang2021unidentified} aims to localize and segment novel objects, but the state-of-the-art model~\cite{maskrcnn} does not perform well as shown in ~\cite{wang2021unidentified}.

We find that the failure of current state-of-the-art models is partly due to the training pipeline, \ie, regarding all regions that are not annotated as the foreground objects as background. 
Even if the background includes \textit{hidden} objects—background objects that are visible but unlabeled—as in Fig.~\ref{fig:motivation}, the models are trained not to detect them, which prevents from learning general objectness. To address this, Kim \etal ~\cite{kim2021learning} proposed to learn the localization quality of region proposals instead of classifying them as foreground vs. background. Their approach samples object proposals close to the ground truth and learns to estimate the corresponding localization quality. While partially mitigating the issue, this approach still needs to carefully set the overlap threshold for positive/negative sampling and risks suppressing hidden objects as non-objects. 

To improve open-world instance segmentation, we propose a simple, yet powerful, learning framework along with a new data augmentation method, called \textit{Learning to Detect Every Thing (\ours )}. To eliminate the risk of suppressing hidden objects, we copy foreground objects using their mask annotation and paste them onto a background image. The background image is synthesized by resizing a cropped patch. By keeping the cropped patch small, we make it unlikely that the resulting synthesized images contain any hidden objects. However, this background creation process makes synthesized images look very different from real images, \eg, the background may consist only of low-frequency content. Thus, a detector naively trained on such images performs poorly. 
To overcome this limitation, we decouple the training into two parts: 1) training background and foreground region classification and localization heads with synthesized images, and 2) learning a mask head with real images. We show that such hybrid training on both domains but with a shared backbone makes the model invariant to the domain shift between augmented and real images.

\ours demonstrates remarkable gains in open-world instance segmentation and detection. On COCO~\cite{coco}, \ours trained on VOC categories improves the average recall by 12.8 points when evaluated on non-VOC categories. Surprisingly, \ours achieves significant improvements in detecting novel objects without requiring additional annotation \eg, \ours trained only on VOC categories (20 classes) in COCO  outperforms Mask RCNN trained on all COCO categories (80 classes) when evaluating average recall on UVO~\cite{wang2021unidentified}. As shown in Fig.~\ref{fig:examples_result_uvo}, \ours can generate precise object proposals as well as cover many objects in the scene.

Our contributions are summarized as follows:
\begin{itemize}
    \item We propose a simple framework, \ours,  consisting of new data augmentation and decoupled training for open-world instance segmentation, which is applicable to both one-stage and two-stage detectors.
    \item We demonstrate that both our data augmentation and decoupled training are crucial to achieving good performance in open-world instance segmentation. 
    \item \ours outperforms state-of-the-art methods in all settings including cross-category settings on COCO and cross-dataset setting on COCO-to-UVO, COCO-to-Object365, and Cityscapes-to-Mapillary. 
\end{itemize}

\vspace{-4mm}
\section{Related Work}
\vspace{-1mm}
\noindent\textbf{Region proposals.}
Unsupervised region proposal generation used to be a standard approach to localize objects in a scene~\cite{zitnick2014edge,alexe2012measuring,arbelaez2014multiscale,uijlings2013selective}. These approaches localize objects in a class-agnostic way, but employ hand-crafted features (\ie., color contrast, edge, \etc) to capture general objectness. 

\noindent\textbf{Closed-World object detection.}
Much effort has been spent on supervised object detection with a closed world assumption~\cite{felzenszwalb2010cascade,girshick2014rich,faster,girshick2015fast,ssd,yolo}. 
The ability to detect known objects has been improving with better architecture designs~\cite{lin2017feature,cai2018cascade,dai2017deformable} or objectives~\cite{lin2017focal}. 
Also, localizing objects given a few training examples or semantic information is becoming a popular research topic~\cite{kang2019few,bansal2018zero}. However, these attempts are still constrained by the taxonomy defined by the dataset. Our model can detect more categories than defined by the dataset, which can be very useful in few-shot or zero-shot object detection. 

\noindent\textbf{Open-World object detection/segmentation.}
Open-world recognition problems are gaining attention in image classification, object detection, and segmentation~\cite{bendale2016towards,dhamija2020overlooked,cen2021openseg}. 
Especially, many methods have been proposed for open-set image classification, where the goal is to separate novel categories from known categories given a closed-set training set~\cite{bendale2016towards,liu2019large,perera2020generative,yoshihashi2019classification,tack2020csi}. On the contrary, the goal of open-world instance segmentation is to detect and segment all objects in a scene without distinguishing novel objects from seen ones. We acknowledge that there is ambiguity in the definition of “object”, and follow \cite{wang2021unidentified} during evaluation.

Wang \etal~\cite{wang2021unidentified} recently published the first benchmark dataset for open-set instance segmentation, which includes various categories from YouTube videos. However, from a methodological perspective, open-world object detection and segmentation remain understudied despite the importance of the task. 
Hu \etal and Kuo \etal~\cite{hu2018learning,kuo2019shapemask} proposed approaches for predicting masks of various objects, but they require bounding boxes from classes of interest. 
Jaiswal \etal~\cite{jaiswal2021class} trained a detector in an adversarial manner to learn class-invariant objectness.
Joseph \etal~\cite{joseph2021towards} proposed a semi-supervised learning approach for open-world detection, which regards regions that are far from ground truth boxes but have a high objectness score as hidden foreground objects. 
Kim~\etal\cite{kim2021learning} employed localization quality estimation with the claim that the estimation strategy is more generalizable in open-world instance segmentation. Note that this is a concurrent work. 

The core of the open-world detection problem lies in the detector training pipeline: regarding hidden objects as background. This training scheme is common in both two-stage and one-stage detectors. However, none of the approaches listed above solves this issue. Our approach takes the first step in addressing background suppression via novel data augmentation strategies and shows remarkable improvements over baselines despite its simplicity. 

\noindent\textbf{Copy-Paste augmentation.}
Pasting foreground objects on a background is a widely used technique in many vision applications~\cite{dosovitskiy2015flownet,ghiasi2021simple,varol17_surreal}.
Recently, copy-and-paste augmentation was shown to be a very useful technique in instance segmentation~\cite{dvornik2018modeling,dwibedi2017cut,ghiasi2021simple}. %These approaches paste objects on other images and train a model on the augmented images. %Such an augmentation procedure is object-aware, both in terms of category and shape, and useful for instance segmentation.
Dwibedi~\etal~\cite{dwibedi2017cut} proposed to synthesize an instance segmentation dataset by pasting object instances on diverse backgrounds and trained on the augmented images in addition to the original dataset. Dvornik~\etal~\cite{dvornik2018modeling} considered modeling the visual context to paste the objects while Ghiasi~\etal~\cite{ghiasi2021simple} showed that pasting objects randomly is good enough to provide solid gains. These methods still assume a closed-world setting, whereas our task is the open-world instance segmentation problem.
There are two technical differences compared to these methods. 
First, our augmentation samples background images from a small region of an original image to create a background unlikely to have any objects. This pipeline is designed to circumvent suppressing hidden objects as background and does not require any external background data as used in \cite{dwibedi2017cut}. 
Second, we decouple the training into two parts, which is also key to achieving a well-performing open-world detection model. In contrast, all of the existing approaches above simply train on synthesized images.

\begin{figure*}[t]
    \centering
    \includegraphics[width=0.9\linewidth]{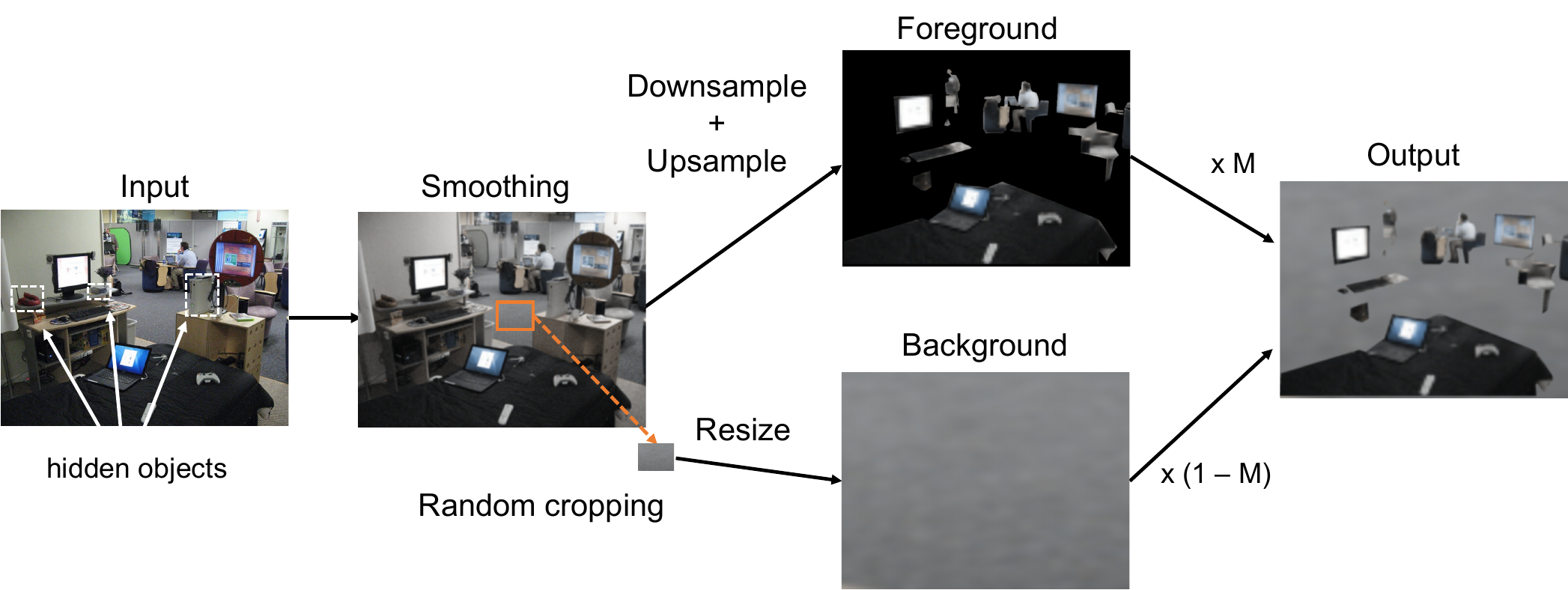}
    \vspace{-2mm}
    \caption{\small{\textbf{Our augmentation strategy} creates images without hidden objects by upscaling small regions to use as background.}}
    \label{fig:pipeline_synthesis}
\end{figure*}
\vspace{-1mm}
\section{Learning to Detect Every Thing}
\vspace{-2mm}
In this section, we describe the proposed \ours scheme for open-world instance segmentation. 
During training, we are given an instance segmentation dataset with annotations of known classes. In testing, the model is required to locate objects of unknown classes. 

Mask-RCNN~\cite{maskrcnn} serves as the base model, but our method is applicable to different architectures such as RetinaNet~\cite{lin2017focal} and TensorMask~\cite{chen_2019_iccv}. We describe details of the data generation process (Fig.~\ref{fig:pipeline_synthesis}) and training scheme (Fig.~\ref{fig:training_figure}) below. 

\iffalse
Our model is trained via the instance mask loss on real images as well as the detector loss on images with synthesized background. We describe details of the data generation process (Fig.~\ref{fig:pipeline_synthesis}) and training scheme (Fig.~\ref{fig:training_figure}) below.
\fi

\vspace{-2mm}
\subsection{Data Augmentation: Background Erasing (BackErase)}
\vspace{-1mm}
We propose a new data augmentation to mitigate the bias induced by unlabeled objects prevalent in most training sets. These hidden objects are not given annotation because they do not belong to known classes or are overlooked by annotators.
We propose to synthesize fully labeled training images. Using the instance mask, we crop only the annotated foreground regions and paste them on the synthesized background canvas. These synthesized images have fully labeled objects and lead to objectness detectors that generalize better to open world settings.

\noindent\textbf{Background region sampling.} 
First, we apply Gaussian smoothing to the input image before cropping the foreground and background region, and denote the smoothed image as $I_1$. By smoothing the whole image before this operation, we expect to reduce the discrepancy in high-frequency content between the foreground and background images. 

\begin{figure}[t]
    \centering
    \includegraphics[width=0.9\linewidth]{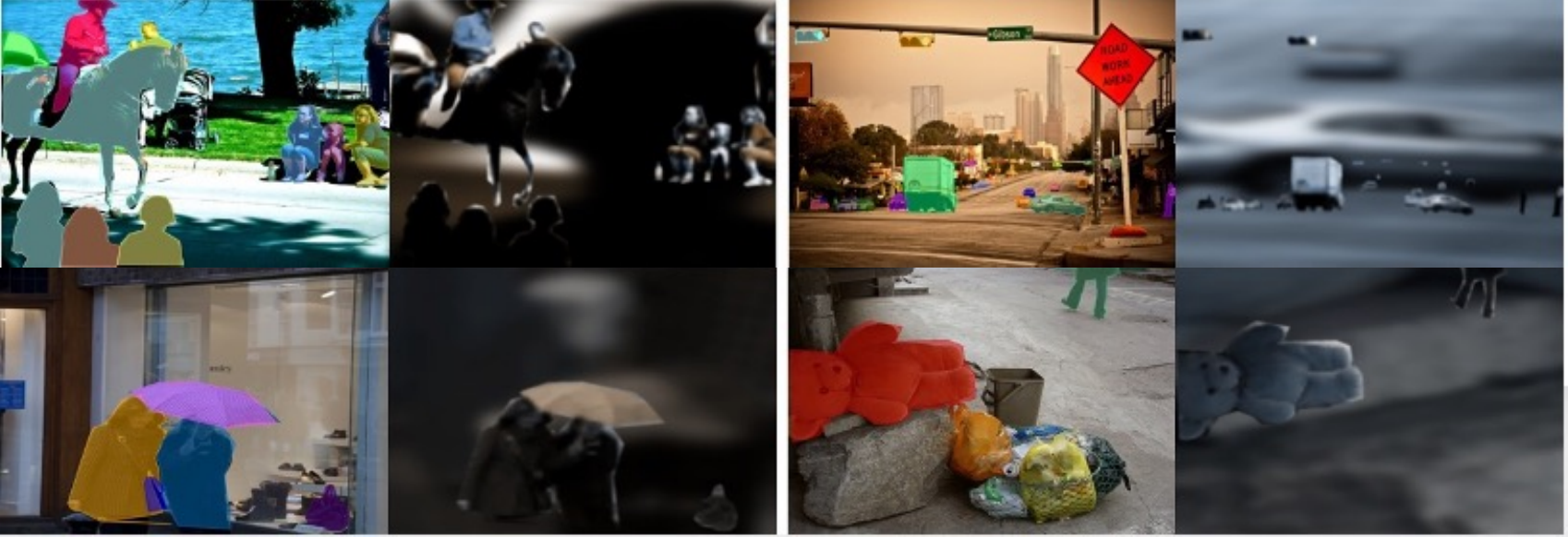}
    \vspace{-3mm}
    \caption{\small{\textbf{Examples of original inputs (odd columns) and synthesized image (even columns).} Masked regions are highlighed with colors (odd columns). Using small regions as background avoids the risk of having hidden objects in the background. We achieve to suppress unlabeled objects present in the background.}}
    \label{fig:examples} \vspace{-3mm}
\end{figure}

Then, we randomly crop a small region from $I_1$, where width and height of the region is set as $\frac{1}{8}$ of the original image's. We resize it to the same size as the input image to serve as a background canvas, which we denote as $I_2$. Cropping a small region entails a much lower risk of including hidden background objects compared to using the original background. Even if it happens to include unannotated objects, drastically upscaling the patch makes the objects' appearance very different, as shown in examples in Fig.~\ref{fig:examples}. We vary the scale of the background canvas in experiments (See Table~\ref{tab:region_size}).

\noindent\textbf{Blending pasted objects.}
To avoid the model learning to separate background and foreground by the difference in frequency information, foreground objects are downsampled and resized to the original size. Then, the foreground objects are pasted on the canvas. 
To insert copied objects into an image, we use the binary mask ($M$) of pasted objects using ground-truth annotations and compute the new image
as $I_1 \times M + I_2 \times (1 - M)$. We apply a Gaussian filter to the binary mask to smooth the edges of the copied objects. 
Examples of synthesized images using the COCO dataset with 80 categories are illustrated in Fig.~\ref{fig:examples}. Note that even in datasets with dense annotations like COCO, many objects are not annotated, and our augmentation effectively removes such hidden objects from the background.
We do not claim that details such as smoothing and resizing operations are necessarily optimal for open-world instance segmentation, but empirically find they work well.

\subsection{Decoupled Multi-Domain Training}
\begin{figure}[t]
    \centering
    \includegraphics[width=\linewidth]{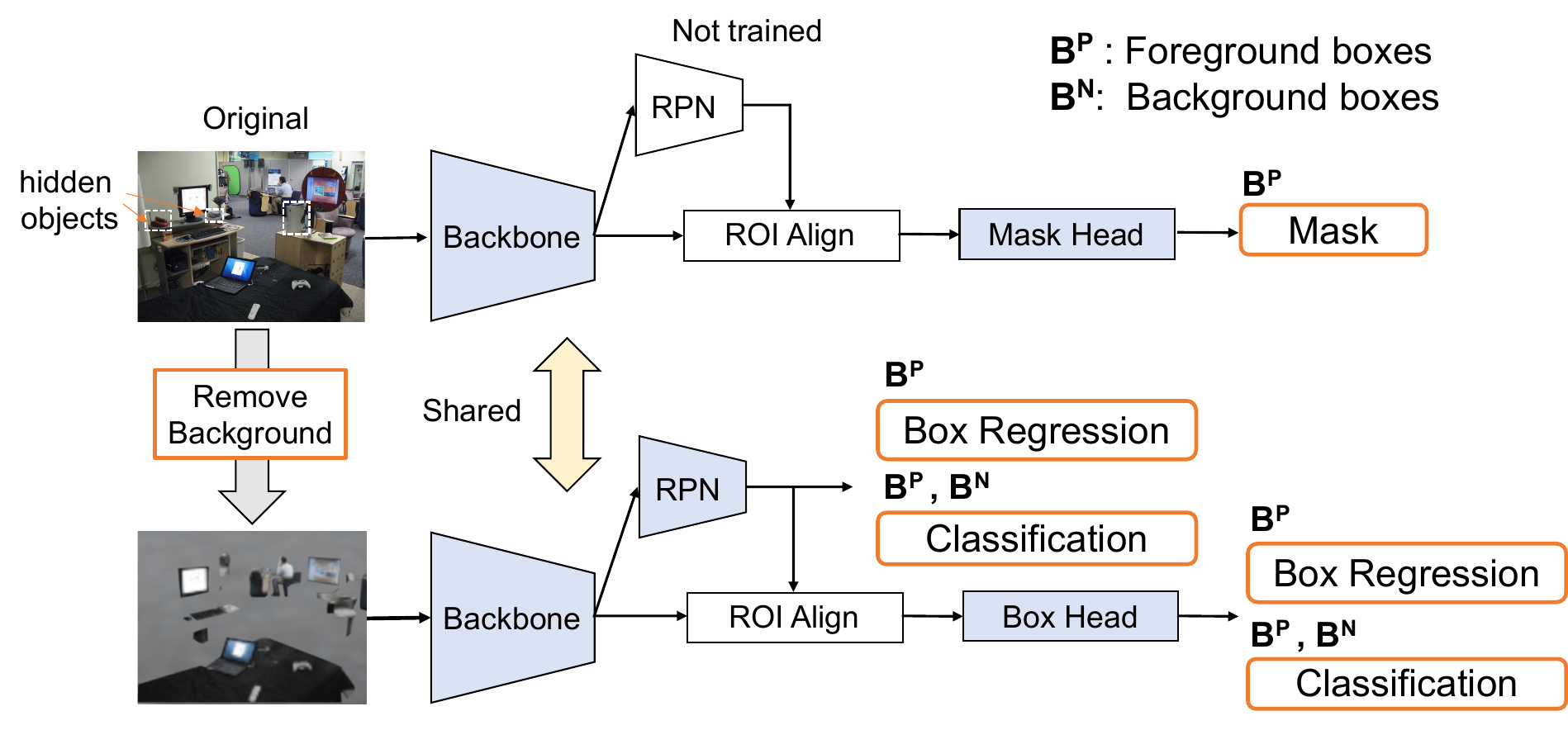}
    \vspace{-5mm}
    \caption{\small{\textbf{Training pipeline.} Given an original input image and synthesized image, we train the detector on the mask loss computed on the original image, and classification, regression losses on the synthesized image.}}
    \label{fig:training_figure}
\end{figure}

Simply training a detector on the synthesized images in the conventional way~\cite{maskrcnn} does not work well due to the domain shift (See Table~\ref{tab:ablation}).
Since real images and our synthesized images have very distinct content and layout, a detector trained on our synthesized data does not generalize well to real images. In this section, we propose a simple yet effective approach to mitigate this issue. 
We solve the shift by computing mask loss on real images while calculating detection loss on synthesized images. Because the backbone is trained on both the synthetic and real domains, it learns an invariance between real and augmented object regions. Even though the losses are different for the two domains, they are highly correlated, which makes the backbone network adapted to real images on both tasks. The entire training pipeline is summarized in Fig.~\ref{fig:training_figure}.

Typically, the training objectives for instance segmentation models consist of two major terms: object detection loss and instance mask loss. 
In methods like Mask RCNN~\cite{maskrcnn}, the object detection loss is composed of a region proposal classification loss and a box regression loss, which are used to train both the region proposal network (RPN) and the region of interest (ROI) heads.
For simplicity, we unify the objectives for RPN and ROI as one loss. 

Each predicted box $B_i$ includes predictions for a box location $\hat{t}_i$, objectness score $\hat{y}_i$, and mask prediction $\hat{m}_i$. During training, $B_i$ comes with corresponding ground truth, $t_i$, $y_i$, and $m_i$. Let $B^{p}$ and $B^{n}$, the positive (foreground) and negative (background) boxes, denote a set of boxes with $y = 1$ and  $y = 0$ respectively. The label of positive or negative, $y$, is decided based on the Intersection over Union (IoU) with the nearest bounding box during training, \eg, regions with IoU smaller/larger than $0.5$ are background/foreground respectively in the case of ROI head. 
Note that, in general, the proposal classification loss is computed for both positive and negative boxes, $B = B^{p} \cup {B^{n}}$, while the box regression and mask loss are computed for ${B^{p}}$.

We highlight that $B^{n}$ from synthetic images ($I_{S}$) are unlikely to contain unlabeled objects due to our augmentation while unlabeled objects can be present in those from real images ($I_{R}$). Given this fact, the detection loss should be computed on boxes of $I_{S}$. However, training only on $I_{S}$ will not make the model generalizable to $I_{R}$ since $I_{R}$ and $I_{S}$ look very different.
Then, to bridge the domain gap between $I_{S}$ and $I_{R}$, we propose to compute the instance mask loss on $I_{R}$. Why does the mask loss help to mitigate the gap \wrt the detection task? Mask prediction aims to separate background and foreground pixels within a foreground bounding box whereas the box classification decides the objectness for one box. The two tasks are very similar in that both attempt to separate background and foreground samples except that the mask loss is computed only for $B^{p}$. Then, the mask loss training signal, which should be useful to solve the detection task for $I_{R}$, is propagated to a deep backbone network shared among the region proposal network, bounding box head, and mask head. The features obtained from the backbone will improve the performance of the box head in $I_{R}$. Furthermore, the model will not learn to suppress unlabeled objects by using the mask loss because the loss is computed on foreground boxes only. In summary, the use of mask loss on real images will make the backbone network adapted to real images without suppressing unlabeled objects. 

Specifically, the loss is computed as follows:
\begin{equation}
\sum_{B_i \in {B^{p}_{aug}}}{\small{L_{reg}(\hat{t}_i, t_i)}} + \sum_{B_i \in B_{aug}} L_{cls}(\hat{y}_i, y_i) + \sum_{B_i \in {B^{p}_{real}}} L_{mask}(\hat{m}_i, {m_{i}})
\end{equation} 
where $L_{reg}$, $L_{cls}$, and $L_{mask}$ indicate the regression, object classification, and mask loss respectively. Note that $B_{aug}$ and $B_{real}$ are used to differentiate the boxes from synthetic and real images.

\noindent\textbf{Class agnostic inference.} 
Since our goal is to detect objects in a scene without classifying them into closed-set classes, class agnostic inference is preferred. We apply a class agnostic inference method to a class discriminative object detector. Given the classification output of a region, we sum up all scores of (known) foreground classes, deeming the result as an objectness score. Mask and box regression are performed for the class with the maximum score.

\vspace{-2mm}
\section{Experiments}
\vspace{-2mm}
We evaluate \ours on two settings of open-world instance segmentation: \textit{Cross-category} and \textit{Cross-dataset}. 
The Cross-category setting is based on the COCO~\cite{coco} dataset, where we split annotated classes into known and unknown classes, train models on known ones, and evaluate detection/segmentation performance on unknown and all classes separately.
Since the model can be exposed to a new environment and encounter novel instances, the Cross-dataset setting evaluates models' ability to generalize to new datasets. For this purpose, we adopt either COCO~\cite{coco} or Cityscapes~\cite{cordts2016cityscapes} as a training source, with UVO~\cite{wang2021unidentified},  Obj365~\cite{shao2019objects365} and Mappilary Vista~\cite{neuhold2017mapillary} as the test datasets.

%%%%%%%%%%%%%%% Train on VOC, test on Non-VOC
\begin{table*}[t]
\centering
\resizebox{0.9\linewidth}{!}{
\begin{tabular}{c|ccc|ccc||cc|cc}
\toprule
\multirow{3}{*}{Method}&  \multicolumn{6}{c||}{Non-VOC}&\multicolumn{4}{c}{All}\\%\cline{2-9}
 & \multicolumn{3}{c|}{\small{Box}}& \multicolumn{3}{c||}{\small{Mask}} & \multicolumn{2}{c|}{\small{Box}}& \multicolumn{2}{c}{\small{Mask}}\\
& \small{AP} & \small{AR$_{10}$}  & \small{AR$_{100}$} & \small{AP} & \small{AR$_{10}$} & \small{AR$_{100}$}&  \small{AR$_{10}$}  & \small{AR$_{100}$} & \small{AR$_{10}$} & \small{AR$_{100}$}\\
\hline 

Mask RCNN~\cite{maskrcnn} &1.5&8.8&10.9&0.7&7.2&9.1&19.3&23.1&16.7&19.9\\
$\text{Mask RCNN}^{\text{P}}$ &1.1&8.7&10.7&0.6&7.2&8.9&19.1&23.0&16.5&19.8\\
$\text{Mask RCNN}^{\text{S}}$ &3.4&13.2&18.0&2.2&11.3&15.8&21.7&27.4&19.2&24.4\\
%10.2	21.5	28.5	34.8		9	19.7	25.6	31

\ours &\textbf{5.0}&\textbf{18.2}&\textbf{30.8}&\textbf{4.7}&\textbf{16.3}&\textbf{27.4}&\textbf{24.4}&\textbf{36.8}&\textbf{22.4}&\textbf{33.1}\\
\bottomrule
\end{tabular}}

\caption{\small{\textbf{Results of {VOC} $\rightarrow$ {COCO} generalization}. \ours outperforms all baselines and showing large improvements on Mask RCNN.}}
\label{tab:voc2nonvoc}
\end{table*}

\begin{figure*}[t]
    \centering
    \includegraphics[width=\linewidth]{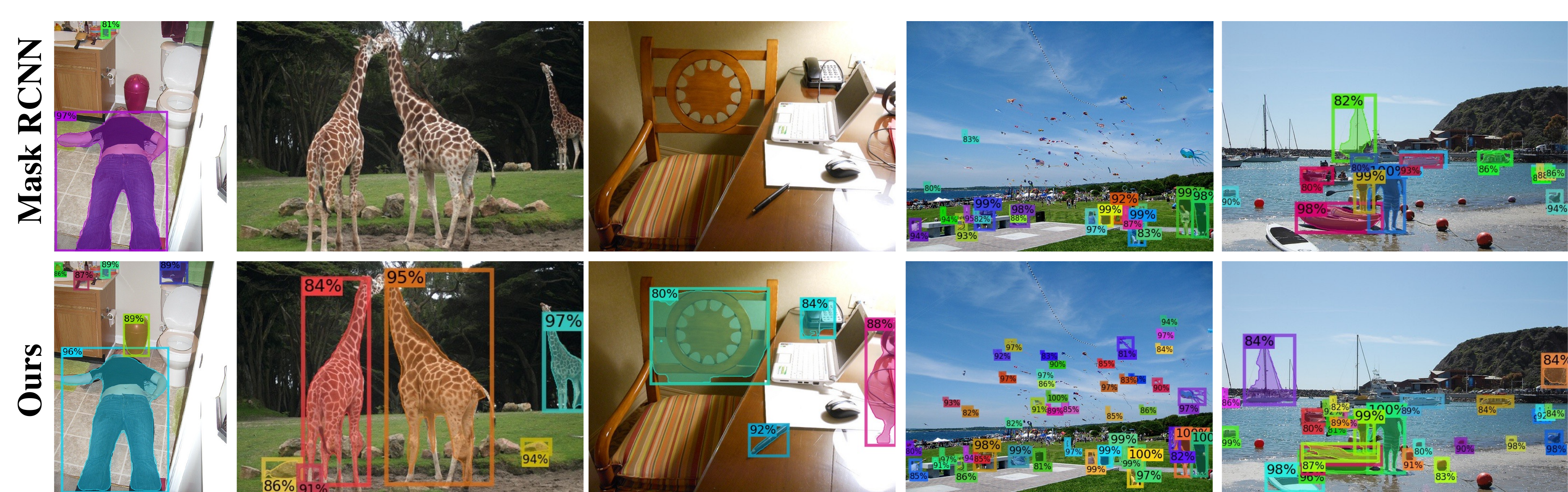}
    \vspace{-6mm}
    \caption{\small{\textbf{Visualization in VOC to Non-VOC in COCO dataset. Top: Mask RCNN. Bottom: \ours.} Note that training categories do not include giraffe, trash box, pen, kite, and floats. \ours detects many novel objects better than Mask RCNN.}}
    \label{fig:examples_result}
\end{figure*}
%%%%%%%%%%%%%%% Train on 
\begin{table}[t]
\small

\centering
%\resizebox{\linewidth}{!}{
\resizebox{0.9\linewidth}{!}{

\begin{tabular}{c|ccccc|ccccc}
\toprule
\multirow{2}{*}{Method}&\multicolumn{5}{c|}{Top-5 }&\multicolumn{5}{c}{Worst-5}\\
&  \small{bear}  & \small{bed}  &  \small{microwave}& {elephant} & \small{t-bear}  & \small{carrot}  & \small{tie}  &  \small{skis}&\small{broccoli} &\small{donut} \\
						
\hline 
Mask RCNN &\textbf{78.6}& 45.2&36.5&28.6&20.3&0.4&0.4&1.2&1.2&1.2\\
\ours &76.5&\textbf{57.6}&\textbf{59.5}&\textbf{67.2}&\textbf{45.9}&\textbf{6.2}&\textbf{1.9}&\textbf{8.1}&\textbf{8.3}&\textbf{15.7}\\
\bottomrule
\end{tabular}}
\caption{\small{\textbf{AR on top- and worst-5 classes detected by Mask RCNN baseline in {VOC} $\rightarrow$ {Non-VOC}.}}
\label{tab:classwise}}
\end{table}

%%%%%%%%%%%%%%% Train on VOC, test on Non-VOC
\begin{table}[t]
\centering
\resizebox{0.7\linewidth}{!}{
\begin{tabular}{c|cc|cc|ccc}
\toprule
\multirow{2}{*}{Method}&\multicolumn{2}{c|}{\small{Box}}&\multicolumn{2}{c|}{\small{Mask}} &\multicolumn{3}{c}{\small{Box}}\\
&\multirow{1}{*}{\small{Real}}&\multirow{1}{*}{\small{Synth}} &\multirow{1}{*}{\small{Real}}&\multirow{1}{*}{\small{Synth}}
& \small{AR$_{10}$}   & \small{AR$_{100}$} &\small{AR0.5} \\
\hline 
Plain&\checkmark & &\checkmark&&8.8&10.9&19.1\\
Synth Only&&\checkmark&&&1.6&4.3&11.7\\
Synth Only$^{*}$&&\checkmark&&\checkmark&3.0&9.5&23.8\\
LDET&&\checkmark&\checkmark&&\textbf{18.2}&\textbf{30.8}&\textbf{53.2}\\
\bottomrule

\end{tabular}}
\caption{\small{\textbf{Ablation study of data and training method in {VOC} $\rightarrow$ {Non-VOC}.} We change the data used to compute detection and mask loss. Training only on synthetic data does not perform well while LDET, which is trained on both data with decoupled training, performs the best.}}
\label{tab:ablation}\vspace{-3mm}
\end{table}

\noindent\textbf{Implementation.}
Detectron2~\cite{wu2019detectron2} is used to implement \ours. 
Mask R-CNN~\cite{maskrcnn} with ResNet-50~\cite{he2016deep} as feature pyramid~\cite{lin2017feature} is used unless otherwise specified. Following~\cite{wang2021unidentified}, we utilize the standard hyperparameters of Mask R-CNN~\cite{maskrcnn} as defined in Detectron2. See appendix for more details \eg, hyper-parameters. 
We will release the full code to reproduce our results upon acceptance.

\noindent\textbf{Baselines.}
Since open-world instance segmentation is a new task, we develop several baselines as follows. See appendix for more details of baselines.

\textit{1) Mask R-CNN.} We adopt the default model without changing any objectives or input data. Comparison to this baseline will reveal the difference from standard training.

\textit{2) $\text{Mask RCNN}^{\text{S}}$.} We avoid sampling background regions with hidden objects by sampling background boxes from the regions mostly inside the ground truth boxes. 
We assume that these regions are less likely to contain hidden objects. We compute the area of intersection with ground truth boxes over the area of the proposal box and sample background boxes with a large value of this criterion. 

\textit{3) $\text{Mask RCNN}^{\text{P}}$.} Inspired by ~\cite{joseph2021towards}, we implement a pseudo-labeling based open-set instance segmentation baseline. The idea is to assign pseudo-labels of foreground classes to the background regions (IoU with GT $<$ 0.5) that have high objectness scores. A model is trained to minimize the box classification loss on pseudo-labels.

\noindent\textbf{Evaluation.}
In this work, Average Recall (AR) is mainly employed for performance evaluation following ~\cite{wang2021unidentified}. When class labels are available, we compute AR for each class given the objectness score and average over all classes as done in the standard COCO evaluation protocol. Unless otherwise specified, AR is computed following the COCO evaluation protocol, \ie, AR at 100 detections. Average precision (AP) is computed in a class agnostic way.

\vspace{-2mm}
\subsection{Cross-category generalization}
\vspace{-2mm}
\noindent\textbf{Setup.}
We split the COCO dataset into 20 seen (VOC) classes and 60 unseen (non-VOC) classes. We train a model only on the annotation of 20 VOC classes. 
The hyper-parameters of baselines and \ours are chosen based on the performance of the randomly selected 20 Non-VOC classes. Then, the whole validation split of COCO is adopted for evaluation. To better understand the results, we report AR on two settings \ie, 60 non-VOC classes only (novel class evaluation) and all 80 classes (generalized evaluation). In evaluating AR in novel class evaluation, we do not count the ``seen class'' detection boxes into the budget of the recall when computing the score. This is to avoid evaluating any recall on seen-class objects. 

\noindent\textbf{Comparison with baselines.}
As shown in Table~\ref{tab:voc2nonvoc}, our method outperforms baselines in all metrics with a large margin. The difference is more evident in the results on non-VOC classes. Some visualizations are available in Fig.~\ref{fig:examples_result}.  
Mask RCNN tends to overlook non-VOC class objects even when they are in the dominant and salient region, \eg, \textit{trash can} (leftmost column) and \textit{giraffes} (second leftmost column). On the other hand, \ours generalizes well to novel objects such as \textit{comb}, \textit{towel}, \textit{giraffes}, \textit{pen}, \textit{phone}, \textit{kites}, and \textit{floats}. 
Table~\ref{tab:classwise} describes AR of top- and worst-5 classes in Mask RCNN. Mask RCNN outperforms LDET on \textit{bear} probably because there are several categories similar to \textit{bear}, \eg, \textit{dog} and \textit{horse} or maybe because there are no 'bear' hidden objects. Although both \ours and the baseline do not excel at detecting classes whose appearance is dissimilar to VOC classes, \ours outperforms Mask RCNN.

\begin{figure}[t]
\begin{subfigure}{.5\textwidth}
  \centering
  \includegraphics[width=.7\linewidth]{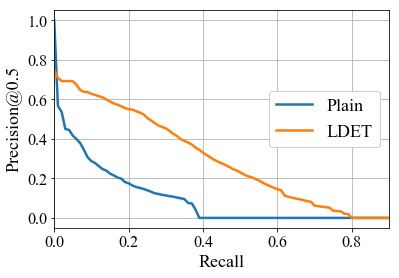}
  \caption{VOC to Non-VOC.}
  \label{fig:voc2nonvoc_pr}
\end{subfigure}%
\begin{subfigure}{.5\textwidth}
  \centering
  \includegraphics[width=.7\linewidth]{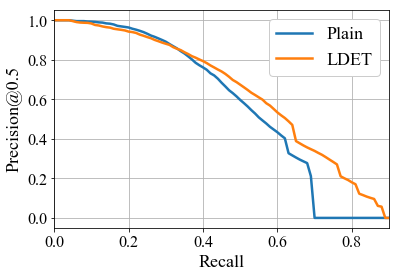}
  \caption{VOC to UVO.}
  \label{fig:voc2uvo_pr}
\end{subfigure}
\vspace{-3mm}
\caption{\textbf{Precision-Recall Curve.} The precision is measured at the IoU threshold of 0.5. The comparison demonstrates that \ours detects novel objects more precisely than the plain model.}
\label{fig:pr_curves}
\end{figure}

\noindent\textbf{Precision-Recall curve.}
Fig.~\ref{fig:pr_curves} (a) shows precision and recall curve measured on non-VOC classes. In most points, the precision of \ours is better than that of the plain model, which means that \ours outputs more precise bounding boxes for novel objects.

\noindent\textbf{Ablation study for learning objectives.}
Table~\ref{tab:ablation} shows an ablation study of training data and objectives. If a model learns only from synthetic data (Synth Only), it fails to detect and segment objects. Interestingly, adding the synthetic mask loss on top of the synthetic detection data (Synth Only$^{*}$) improves the performance. This supports our claim that mask prediction and detection tasks are highly correlated. Adding the mask loss even for synthetic data provides a better understanding of the location of the objects and improves performance. Computing detection loss of synthetic data and mask loss on real data obtains the best results. These results indicate that our proposed decoupled training is very suitable for the open-world instance segmentation and detection task.

\begin{figure}[t]
    \centering
        \vspace{-1mm}
    \includegraphics[width=\linewidth]{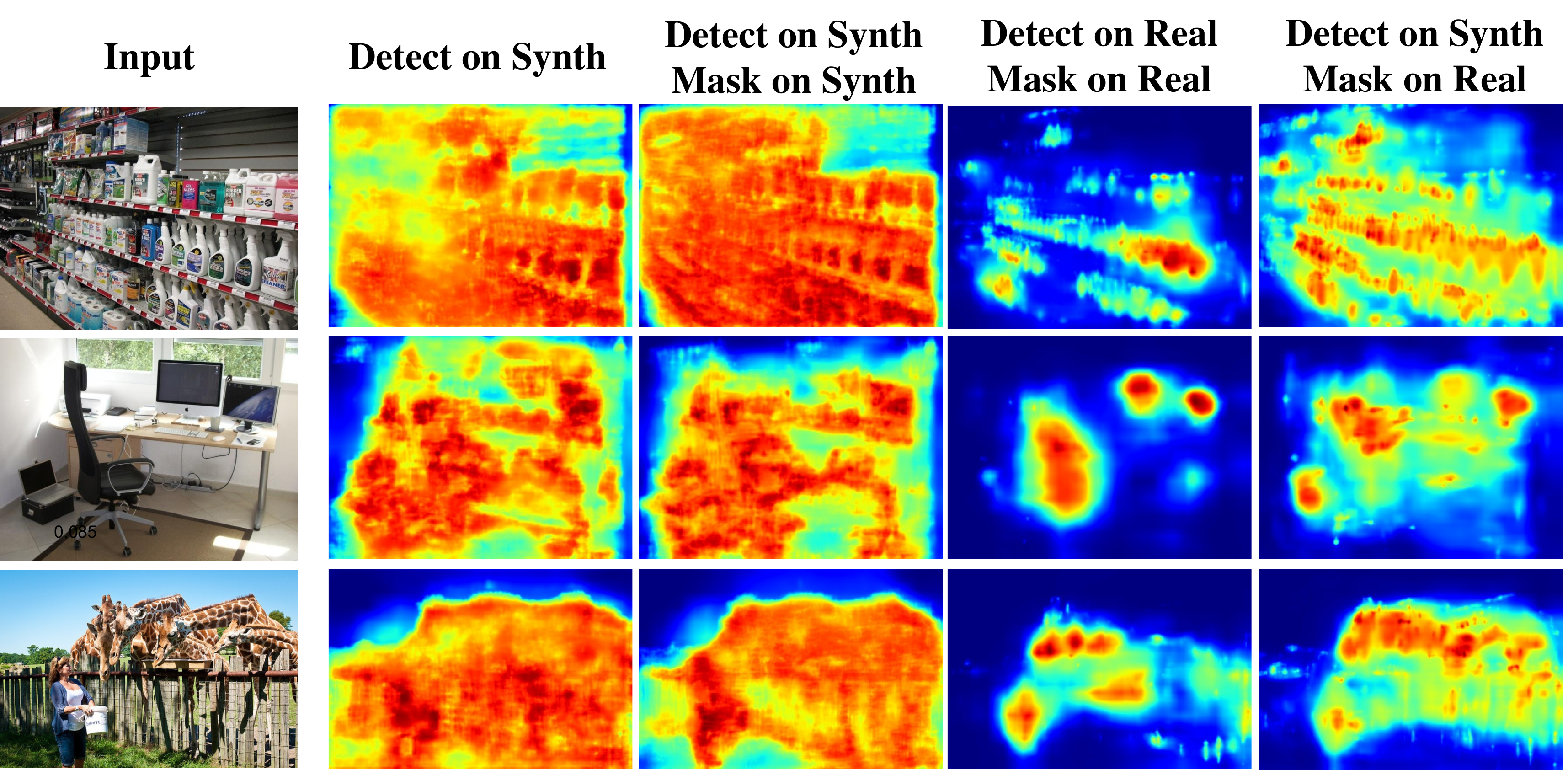}
    \vspace{-7mm}
    \caption{\small{\textbf{Objectness map (RPN score) visualization \wrt different data used to compute detection and mask loss.} A model only with the detection loss for synthetic data (leftmost) does not suppress background regions well. Adding mask loss on real data (rightmost) captures objectness of various categories whereas a plain model (second from the right) tends to suppress many objects.}}
    %\ours (rightmost) captures objectness of various categories whereas Mask RCNN tends to suppress many objects.}}
    \label{fig:silency}
\end{figure}

\noindent\textbf{Visualization of the learnt objectness map.}
Fig.~\ref{fig:silency} visualizes the confidence scores map of the region proposal network, computed by averaging outputs from all feature pyramids. Models trained only on synthetic data (first and second from the leftmost) cannot separate foreground and background well though they seem to cover many foreground objects. While the model trained only with real data (second from rightmost) suppresses the score for many objects, \ours (rightmost) correctly captures objectness for diverse objects. For instance, it captures objectness well in an image crowded with objects in the first row. The mask loss on real images seems to help the detector to separate foreground and background well. 

%%%%%%%%%%%%%%% Train on VOC, test on Non-VOC
\begin{table}[t]
\centering
\vspace{-3mm}
\resizebox{0.7\linewidth}{!}{
\begin{tabular}{c|cccccc}
\toprule
Background ratio  &AP&AR$_{10}$&AR$_{100}$ & AR$_{\text{small}}$ & AR$_{\text{medium}}$ & AR$_{\text{large}}$\\\hline
$2^{-1}$ &\textbf{5.2}&16.8&26.2&17.0&27.0&40.6\\
$2^{-2}$ &5.0&16.5&25.9&17.3&27.1&38.9\\

$2^{-3}$ &5.0&\textbf{18.2}&30.8&18.8&34.6&44.5\\
 $2^{-4}$ &\textbf{5.2}&17.5&\textbf{31.8}&\textbf{20.1}&\textbf{35.8}&\textbf{45.6}\\
\bottomrule
\end{tabular}}
\caption{\small{\textbf{Varying the size of background regions.} $2^{-m}$ indicates cropping background region with $2^{-m}$ of width and height of an input image.
Sampling background from smaller region tends to improve AR.}}
\label{tab:region_size}
\end{table}

%%%%%%%%%%%%%%% Train on VOC, test on Non-VOC
\begin{table}[t]
%\begin{tabular}{ll}
\begin{minipage}[t]{0.43\textwidth}
\centering
%\vspace{-4mm}
\begin{tabular}[t]{c|c|ccccc}
\toprule
\multirow{1}{*}{Method}&\multirow{1}{*}{\small{Detector}} &AR$_{10}$&AR$_{50}$&AR$_{100}$ \\
\hline  
Plain &RPN&\textbf{11.0}&\textbf{19.4}&\textbf{22.9}\\
Plain &ROI&8.8&10.8&10.9\\\hline
\ours&RPN&15.4&	26.4&30.8\\
\ours &ROI&\textbf{18.2}&\textbf{28.0}&\textbf{30.8}\\
\bottomrule
\end{tabular}
\caption{\small{\textbf{Comparison between region proposal network and region of interest head.}}}
\label{tab:rpn_roihead} 
\end{minipage}
\hspace{3mm}
\begin{minipage}[t]{0.4\textwidth}
\centering
\begin{tabular}[t]{c|c|ccccc}
\toprule
\multirow{1}{*}{\small{Detector}}&\multirow{1}{*}{\small{Method}} &AR$_{10}$&AR$_{50}$&AR$_{100}$ \\
\hline  
\multirow{2}{*}{\small{RetinaNet}}&Plain&9.9&15.7&17.8\\
&\ours&\textbf{15.3}&\textbf{26.7}&\textbf{31.0}\\\hline
\multirow{2}{*}{\small{TensorMask}}&Plain&10.6&17.6&19.7\\
&\ours&\textbf{16.3}&\textbf{26.8}&\textbf{31.1}\\
\bottomrule
\end{tabular}
\caption{\small{\textbf{Results on RetinaNet and TensorMask}}}
\label{tab:retina_tmask} 
\end{minipage}
%\caption{\small{\textbf{Results on RetinaNet and TensorMask}}}

\end{table}

\noindent\textbf{Size of background regions.} 
The size of background regions can be important in our data augmentation: if the size is close to an original image, the background will include many hidden objects. We analyze the effect of the region's size in Table~\ref{tab:region_size}, wherein AR gets better with smaller sizes. This indicates that smaller backgrounds prevent sampling hidden objects and coverage of novel objects (AR) becomes better.

\noindent\textbf{External data for background.} 
Using an external background dataset is an alternative way to synthesize background regions although an image without background is not always accessible. In this experiment, we use DTD~\cite{cimpoi2014describing} (texture image dataset). The background region is replaced with a randomly cropped patch from DTD dataset, and a model is trained in the same way as \ours. See the appendix for more details of training. The resulting AR is 26.7 ($\text{\ours}-3.1$) in bounding box localization. The dataset includes a considerable number of objects despite being primarily a texture dataset, which is probably the cause of degradation in AR.

%%%%%%%%%%%%%%% Train on COCO, test on UVO
\begin{table*}[t]
\centering
\small
\resizebox{\linewidth}{!}{
\begin{tabular}{c|c|ccccc|ccccc}
\toprule
\multirow{2}{*}{Method} & \multirow{2}{*}{Train} &\multicolumn{5}{c|}{Box}& \multicolumn{5}{c}{Mask}\\
&& {\small{AP}} & \small{AR}  & \small{AR$_{\text{small}}$}  & \small{AR$_{\text{med}}$}  &  \small{AR$_{\text{large}}$}& {\small{AP}} & \small{AR}  & \small{AR$_{\text{small}}$}  & \small{AR$_{\text{med}}$}  &  \small{AR$_{\text{large}}$} \\
\hline 
Mask RCNN&\multirow{4}{*}{\shortstack{VOC \\ (COCO)}}&19.8&30.0&10.7&21.3&43.0&15.5&23.9&9.2&18.5&32.8\\
$\text{Mask RCNN}^{\text{P}}$&&19.2&30.1&10.6&21.3&43.3&15.4&24.1&9.4&18.4&33.2\\
$\text{Mask RCNN}^{\text{S}}$&&19.7&32.0&10.0&23.3&46.0&14.1&25.9&9.5&20.2&35.4\\
\ours &&\textbf{22.4}&\textbf{43.7}&\textbf{24.7}&\textbf{39.9}&\textbf{52.9}&\textbf{18.4}&\textbf{36.0}&\textbf{22.1}&\textbf{34.8}&\textbf{41.4}\\
\hline
Mask RCNN& \multirow{4}{*}{\shortstack{COCO}}&25.3&42.3&22.2&38.3&52.0&20.6&35.9&19.6&33.9&42.6\\
$\text{Mask RCNN}^{\text{P}}$ &&24.4&41.9&22.3&37.8&51.5&20.1&35.4&19.7&33.6&41.8\\
$\text{Mask RCNN}^{\text{S}}$ &&23.4&40.5&17.6&34.9&52.3&18.0&34.7&16.6&31.5&42.8 \\
\ours & &\textbf{25.8}&\textbf{47.5}&\textbf{29.1}&\textbf{44.8}&\textbf{55.6}&\textbf{21.9}&\textbf{40.7}&\textbf{26.8}&\textbf{40.0}&\textbf{45.7}\\
\bottomrule

\end{tabular}}
\caption{\small{\textbf{Results of {COCO} $\rightarrow$ {UVO} generalization. Top rows: Models trained on VOC-COCO. Bottom rows:  Models trained on COCO.} \ours demonstrates high AP and AR in all cases compared to baselines.}}
\label{tab:coco2uvo}
\end{table*}

%%%%%%%%%%%%%%% Train on VOC, test on Non-VOC
\begin{table*}[t]
\centering
\vspace{-3mm}
\resizebox{0.85\linewidth}{!}{
\begin{tabular}{c|ccccc||cccc}
\toprule
\multirow{2}{*}{Method}&  \multicolumn{5}{c||}{Non-COCO}&\multicolumn{4}{c}{All}\\%\cline{2-9}
&  \small{AP}  &  \small{AR}  & \small{AR$_{\text{small}}$} & \small{AR$_{\text{med}}$} & \small{AR$_{\text{large}}$}&  \small{AR}  &  \small{AR$_{\text{small}}$} & \small{AR$_{\text{med}}$} & \small{AR$_{\text{large}}$}\\\hline 
Mask RCNN~\cite{maskrcnn} &11.9&34.4&21.2&36.0&45.8&38.5&24.0&40.1&50.2\\
$\text{Mask RCNN}^{\text{P}}$ &11.8&32.7&17.5&33.5&47.1&38.6&24.5&39.8&50.8\\
$\text{Mask RCNN}^{\text{S}}$ &10.9&34.6&21.9&35.7&46.6&35.9&18.9&36.6&50.2\\

\ours &\textbf{12.9}&\textbf{38.9}&\textbf{25.5}&\textbf{41.8}&\textbf{50.2}&\textbf{41.1}&\textbf{26.1}&\textbf{43.8}&\textbf{52.8}\\
\bottomrule
\end{tabular}}

\caption{\small{\textbf{Results of {COCO} $\rightarrow$ {Obj365} generalization}. Improvement on Mask RCNN is shown next to each result in the row of \ours. \ours outperforms all baselines and showing large improvements on Mask RCNN.}}
\label{tab:coco2obj365}
\end{table*}

\begin{table}[t]
\centering
%\vspace{-3mm}
\resizebox{0.6\linewidth}{!}{
\begin{tabular}{c|cccc|cccc}
\toprule
\multirow{2}{*}{Method} & \multicolumn{4}{c|}{Box}& \multicolumn{4}{c}{Mask}\\

&  \small{AP}  & \small{AR$_{\text{10}}$}   &  \small{AR$_{\text{100}}$}   &  \small{AR$_{\text{0.5}}$} &  \small{AP}&\small{AR$_{\text{10}}$}   &  \small{AR$_{\text{100}}$}   &  \small{AR$_{\text{0.5}}$}   \\
\hline 
Mask RCNN&8.2&7.7&11.1&20.2&7.3&6.1&8.4&16.3\\
$\text{Mask RCNN}^{\text{P}}$&6.9&7.4&10.8&19.3&7.5&5.5&7.9&16.3\\
$\text{Mask RCNN}^{\text{S}}$ &8.3&6.7&13.3&26.9&6.3&5.5&10.2&21.0\\
\ours &\textbf{8.5}&\textbf{8.0}&\textbf{14.0}&\textbf{28.0}&\textbf{7.8}&\textbf{6.7}&\textbf{10.6}&\textbf{21.8}\\
\bottomrule
\end{tabular}}
\caption{\small{\textbf{Results of {Cityscapes} $\rightarrow$ {Mappilary Vista} generalization.} \ours is effective for autonomous driving dataset. AR$_{\text{0.5}}$ denotes AR with IoU threshold $=0.5$}}
\label{tab:city2map}
\end{table}

\begin{figure*}[t]
    \centering
    \vspace{-6mm}
    \includegraphics[width=\linewidth]{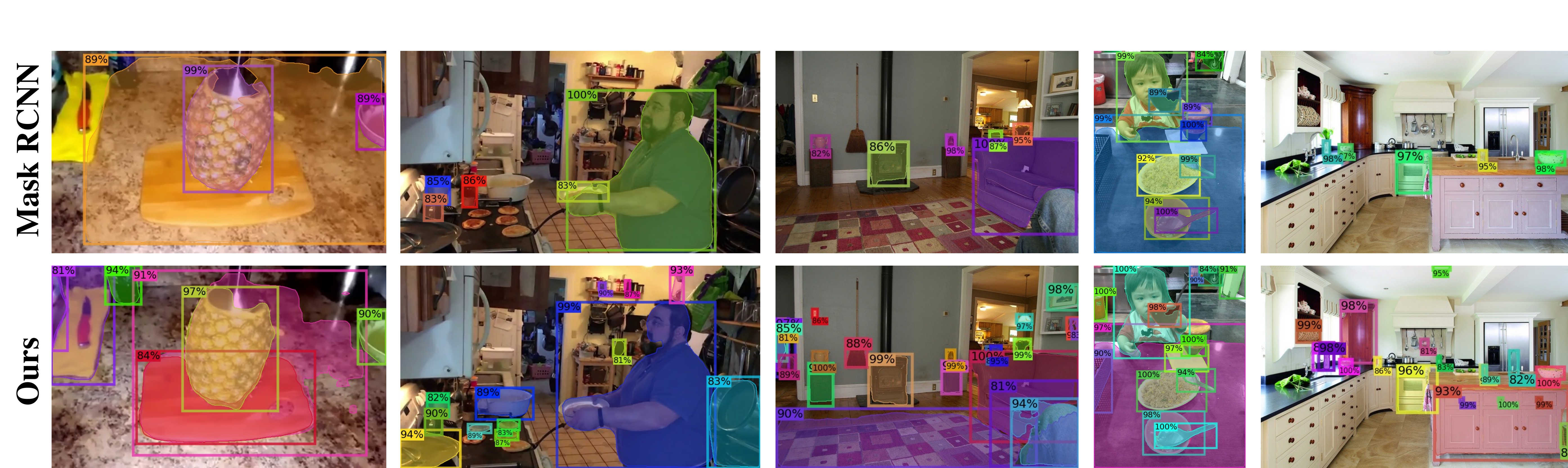}
    \vspace{-6mm}
    \caption{\small{\textbf{Visualization of results for models trained on COCO.} Top: Mask RCNN. Bottom: \ours. Leftmost two images are from UVO, the others are from COCO validation images.}}
    \label{fig:examples_result_uvo_additional}    \vspace{-4mm}

\end{figure*}

\noindent\textbf{Region proposal network (RPN) and region of interest (ROI) head.} 
We compare the performance of RPN and ROI heads in Table~\ref{tab:rpn_roihead}. In Mask RCNN, RPN covers more novel objects, which means ROI learns to suppress many novel objects. In contrast, ROI of \ours is comparable or better than RPN in all metrics.

\noindent\textbf{Evaluation on different architectures.} 
We evaluate \ours on one-stage detectors, RetinaNet~\cite{lin2017focal} and TensorMask~\cite{chen_2019_iccv} in Table~\ref{tab:retina_tmask}. See appendix for the details of the experiment.
\ours shows clear gains over the baseline, which shows that \ours is a universal approach that can be integrated into diverse detectors.

\vspace{-2mm}
\subsection{Cross-dataset generalization}
\vspace{-2mm}
\noindent\textbf{COCO to UVO.}
First, we utilize UVO~\cite{wang2021unidentified}, which covers many categories outside COCO as 57\% object instances do not belong to any of the 80 COCO classes. Since UVO is based on Youtube videos, the appearance is very different from COCO \eg, some videos are egocentric views and have significant motion blur. We test models trained on the COCO-VOC split or the whole COCO. The validation split is used to measure the performance. Since this dataset does not provide class labels, we evaluate the performance in a class-agnostic way.
As shown in Table~\ref{tab:coco2uvo}, in both COCO-VOC and COCO settings, \ours outperforms baselines with a large margin. Note that our VOC-COCO model outperforms Mask RCNN trained on COCO in many metrics. This indicates the remarkable label efficiency of \ours in open-world instance segmentation.
Unlike the result in VOC-NonVOC experiment, the AP of $\text{Mask RCNN}^{\text{S}}$ drops compared to Mask RCNN, probably because their region sampling leads to imbalanced sampling with regions with different scales. Fig.~\ref{fig:pr_curves}(b) describes the trade-off between precision and recall in this setting, which shows the advantage of \ours at most points.

\noindent\textbf{COCO to Obj365.}
Second, we evaluate models on the validation split of Obj365~\cite{shao2019objects365} detection dataset, wherein 60\% object instances do not belong to any of the 80 COCO classes. We test models trained on the whole coco, and evaluation is done in the way as cross-category setting.
As shown in Table~\ref{tab:coco2obj365}, in both non-COCO categories and all categories, \ours outperforms all baselines. This result confirms that \ours is generalizable to detect various categories of objects. 

\noindent\textbf{Cityscape to Mapillary.}
We examine performance in autonomous driving scenes. Detectors are trained on Cityscape~\cite{cordts2016cityscapes} (8 foreground classes, \textit{person, rider, car, truck, bus, train, motorcycle,} and \textit{bicycles}) and tested on the validation set of Mapillary Vistas~\cite{neuhold2017mapillary} with 35 foreground classes including not only vehicles, but also animals, \textit{trash can, mailbox}, and so on.
In Table~\ref{tab:city2map}, \ours shows solid gains over baselines, though this setting is very challenging. Note that the model is trained only on 8 classes and is required to generalize to all the 35 classes in the test data, which explains the lower performance compared to experiments on COCO. 
The result demonstrates that \ours generalizes to datasets other than COCO and can be useful for autonomous driving systems. 

\noindent\textbf{Visualization.} 
As Fig.~\ref{fig:examples_result_uvo} shows, Mask RCNN detector fails in localizing objects unseen in their 80 categories while our detector shows surprisingly good generalization. Note that in the second leftmost row, it recognizes a character drawn on the wall, which is clearly outside COCO categories. Fig.~\ref{fig:examples_result_uvo_additional} visualizes other examples from UVO and COCO.

\vspace{-3mm}
\section{Conclusion}
\vspace{-2mm}
\label{sec:conclusion}
In this paper, we presented a simple approach, \ours,  for the challenging task of open-world instance segmentation. \ours consists of synthesizing images without hidden objects in their background as well as decoupled training for real and synthesized images. \ours demonstrates strong performance on a benchmark dataset of open-world instance segmentation and promising results on autonomous driving datasets. We hope that \ours becomes a simple baseline and accelerates further research in this area.

\noindent\textbf{Limitations.}
As seen in several visualizations, \ours still fails in detecting some novel objects although its performance is much better than baselines. If the appearance of novel objects is distinct from known objects, \ours and most baselines may miss them. 
Also, experiments on Cityscapes (Table~\ref{tab:city2map}) indicate the importance of covering various categories in training data. One way to overcome this limitation is to annotate a wide range of categories for training data. 

\noindent\textbf{Acknowledgments. }This work was supported by DARPA LwLL and NSF Award No. 1535797.  We thank Donghyun Kim and Piotr Teterwak for giving feedback on the draft

% renew command
\setcounter{section}{0}
\setcounter{table}{0}
\setcounter{figure}{0}

\def\thesection{\Alph{section}}
\renewcommand{\thetable}{\Alph{table}}
\renewcommand{\thefigure}{\Alph{figure}}
\appendix
\section{Experimental Details}
In this appendix, we provide experimental details, additional analysis, and visualizations.

\noindent\textbf{Data augmentation.} Alg.~\ref{algo:augmentation} shows the pytorch-style pseudo-code for our data augmentation. 

\noindent\textbf{Implementation.}
We use the default hyper-parameters, provided by Detectron2, to train and test \ours, Mask RCNN, $\text{Mask RCNN}^{\text{S}}$, and $\text{Mask RCNN}^{\text{P}}$. Default configuration files are used for COCO \footnote{\url{detectron2/blob/main/configs/COCO-InstanceSegmentation/mask_rcnn_R_50_FPN_1x.yaml}} and Cityscapes \footnote{\url{detectron2/blob/main/configs/Cityscapes/mask_rcnn_R_50_FPN.yaml}} respectively. 2 GPUs of NVIDIA RTX A6000 with 48GB are used to train models.

\noindent\textbf{One-Stage Detector.}
We use the same hyper-parameters for data augmentation as in Mask RCNN. Also, we follow the default hyper-parameters of RetinaNet and TensorMask. Since RetinaNet does not have mask head by default, we add the mask head on top of the feature pyramid following Mask RCNN. We will publish the code of one-stage detector upon accepntance.

%\noindent\textbf{Test time hyper-parameters.} Following ~\cite{kim2021learning}, we set hyper-parameters in testing as follows: threshold of non-maximum suppression in the roi head is set as 0.7 (0.5 by default) and, in the region proposal network, the number of region proposals kept after the suppression is set as 2000 (1000 by default). Table ~\ref{tab:test_hp} shows the performance difference by the hyper-parameters. AP tends to improve by decreasing the threshold value and degrade by increasing the number of proposals. Low threshold suppresses more boxes, which results in improving AP and degrading AR. Changing the number of proposals has a similar effect. We can observe that the difference is not significant.
%\input{tables/test_time_parameters}

\noindent\textbf{Baselines.}
%We will release the code for these baselines upon acceptance.

\textit{1) Mask R-CNN.} We do not make any change to the default training configuration. 

\textit{2) $\text{Mask RCNN}^{\text{S}}$.} We compute the area of intersection with ground truth boxes over the area of the proposal box, which we call \textit{IoA}, and sample background boxes with a large value of this criterion. In both region proposal network and roi head, we pick background regions whose IoA is larger than 0.7. 

\textit{3) $\text{Mask RCNN}^{\text{P}}$.} Given the classification output (after softmax) from roi head, boxes confidently predicted as one of the foreground classes are choosen from background regions. The threshold to pick the pseudo-foreground is set as 0.9. The classification loss on the pseudo-foreground regions is incorporated to train the detector.

\noindent\textbf{Experiments on texture dataset.} To make a background using images of DTD~\cite{cimpoi2014describing}, we crop the patch with the size of 256 x 256, and rescale it to the size of a detection training image. Then, we blend the foreground and background in the same way as \ours.

\definecolor{commentcolor}{RGB}{110,154,155}   % define comment color
\newcommand{\PyComment}[1]{\ttfamily\textcolor{commentcolor}{\# #1}}  % add a "#" before the input text "#1"
\newcommand{\PyCode}[1]{\ttfamily\textcolor{black}{#1}} % \ttfamily is the code font
\begin{algorithm}[ht]
%\SetAlgoLined
    \PyComment{scale=$\frac{1}{8}$: the size of background region to crop.} \\
    \PyComment{M: mask of the foreground regions.} \\
    \PyComment{Apply gaussian smoothing.} \\
    \PyCode{image = gaussian(image)}\\
    \PyCode{w, h = image.shape}\\
    \PyComment{Randomly crop background with the specified size.}\\ 
    \PyCode{backg =  randomcrop(image, w*scale, h*scale)}\\
    \PyComment{Upscale to the size of the input.}\\ 
    \PyCode{backg = upscale(backg, scale)}\\
    \PyComment{Downsample the input.}\\ 
    \PyCode{image = downsample(image, scale)}\\
    \PyComment{Upsample to the original size.}\\ 
    \PyCode{image = upscale(image, scale)}\\
    \PyComment{Paste foreground objects on the synthesized background.}\\ 
    \PyCode{image = M * image + (1 - M) * backg} \\
    \PyComment{Apply smoothing.} \\
    \PyCode{image = smooth(image)} \\
\caption{PyTorch-style pseudocode for our data augmentation}
\label{algo:augmentation}
\end{algorithm}

\begin{figure*}[t]
    \centering
    \includegraphics[width=\linewidth]{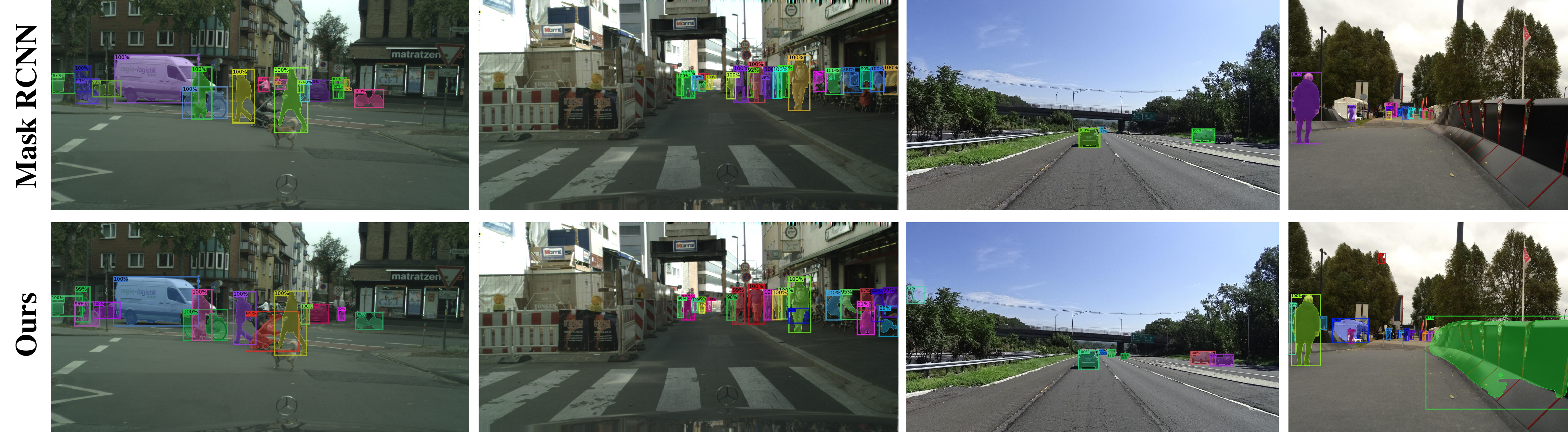}
    \vspace{-6mm}
    \caption{\small{\textbf{Visualization for detectors trained on Cityscapes.} Leftmost two images are validation images of Cityscapes, rightmost two are from Mapillary.}}
    \label{fig:cityscape_examples}
\end{figure*}

\section{Analysis}
\begin{figure}
\begin{subfigure}{.5\textwidth}
  \centering
  \includegraphics[width=.8\linewidth]{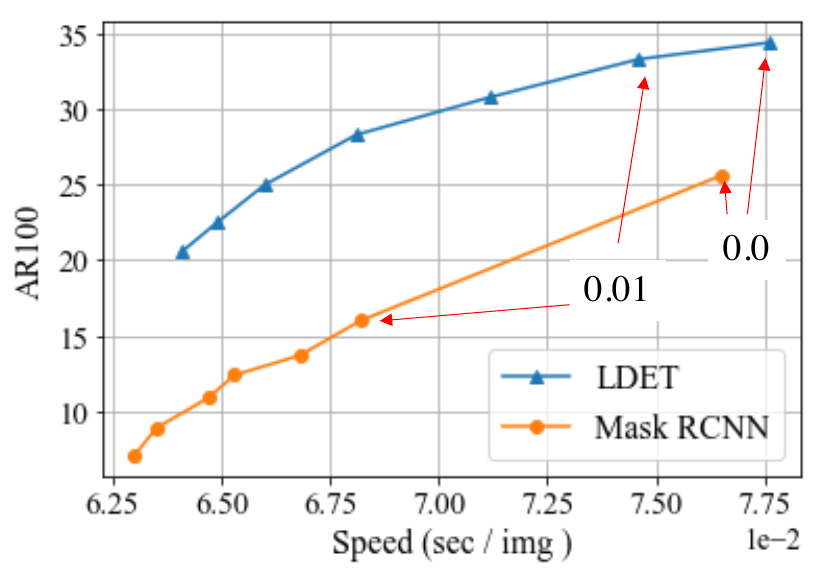}
  \caption{VOC to Non-VOC.}
  \label{fig:voc2nonvoc_speed}
\end{subfigure}%
\begin{subfigure}{.5\textwidth}
  \centering
  \includegraphics[width=.8\linewidth]{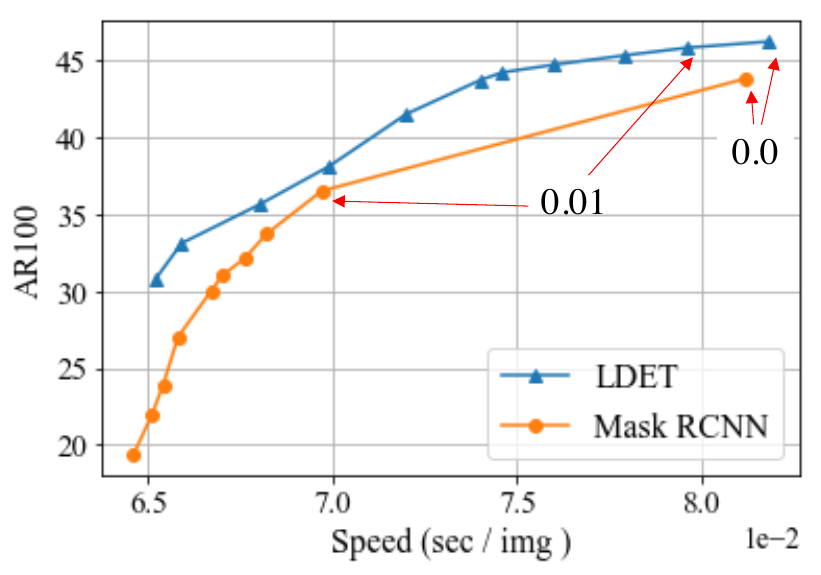}
  \caption{VOC to UVO.}
  \label{fig:voc2uvo_speed}
\end{subfigure}
\vspace{-3mm}
\caption{\textbf{Speed (sec /image) v.s. AR.} We vary the confidence threshold of the ROI head and see the changes of the speed and AR. Note that the speed changes due to the non-maximum supression after confidence thresholding. Points at confidence threshold at 0.0 and 0.01 are highlighted with red arrows. The baseline mask rcnn significantly drops performance between the points at 0.0 and 0.01, which indicates that the model suppresses many foreground objects at the confidence value of 0.01. }
\label{fig:speed_accuracy}
\end{figure}
\noindent\textbf{Study on the confidence threshold.}
In Fig.~\ref{fig:speed_accuracy}, we vary the confidence threshold used to remove unconfident bounding boxes in ROI classification head, where the value is set as 0.05 by default. Here, we vary thresholds starting from 0.0 (no thresholding) to 0.5. This result demonstrates that the baseline drops AR by applying a very small threshold value (Compare AR at 0.0 and 0.01), meaning that the baseline confuses many novel objects with the background.

\begin{figure*}[t]
    \centering
    \includegraphics[width=0.85\linewidth]{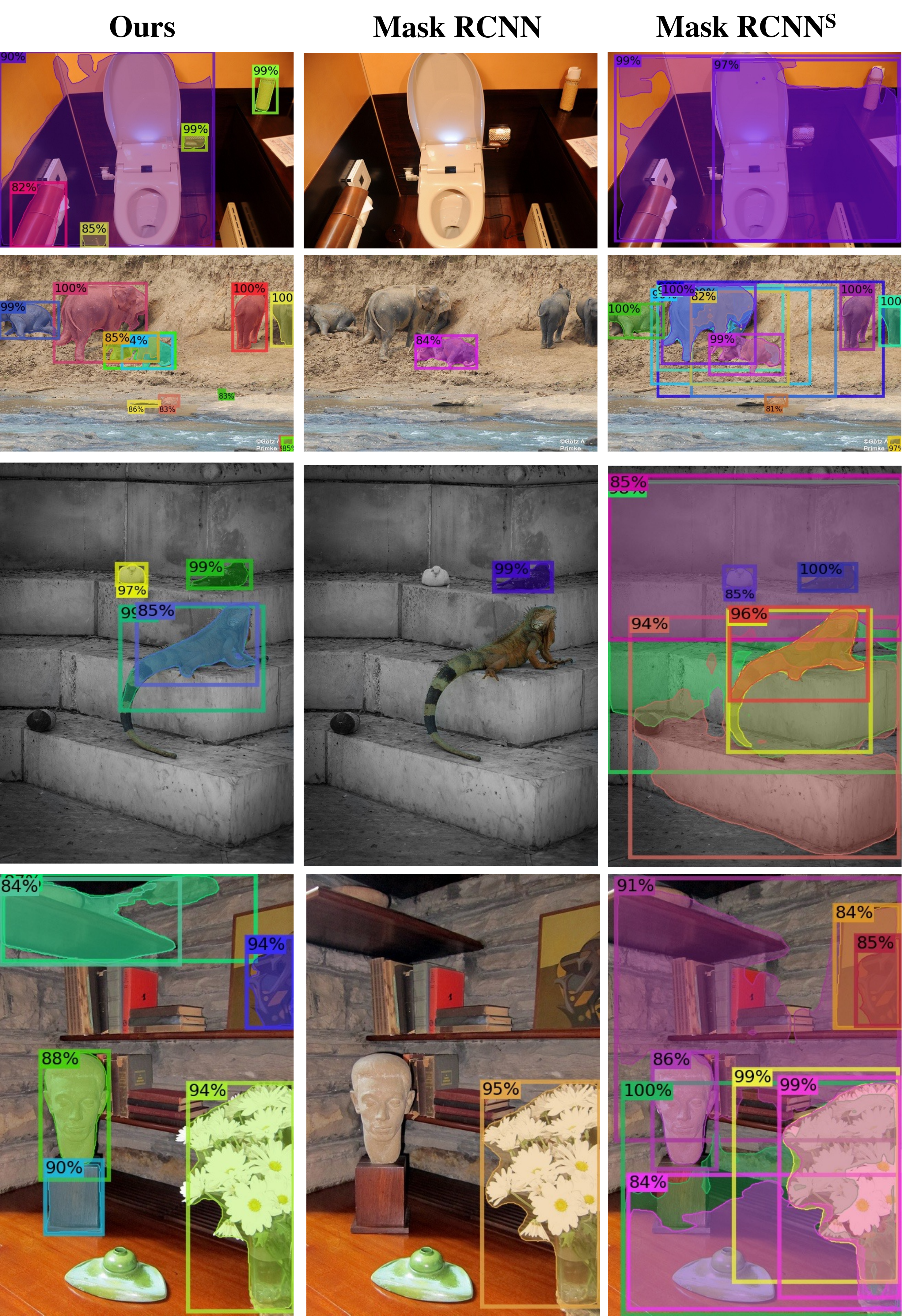}
    \vspace{-3mm}
    \caption{\small{\textbf{Visualization in VOC-COCO to COCO setting.} Note that VOC-COCO does not contain objects such as lizard, toilet paper, and elephant. }}
    \label{fig:voc_to_coco_appendix}
\end{figure*}
\begin{figure*}[t]
    \centering
    \includegraphics[width=0.9\linewidth]{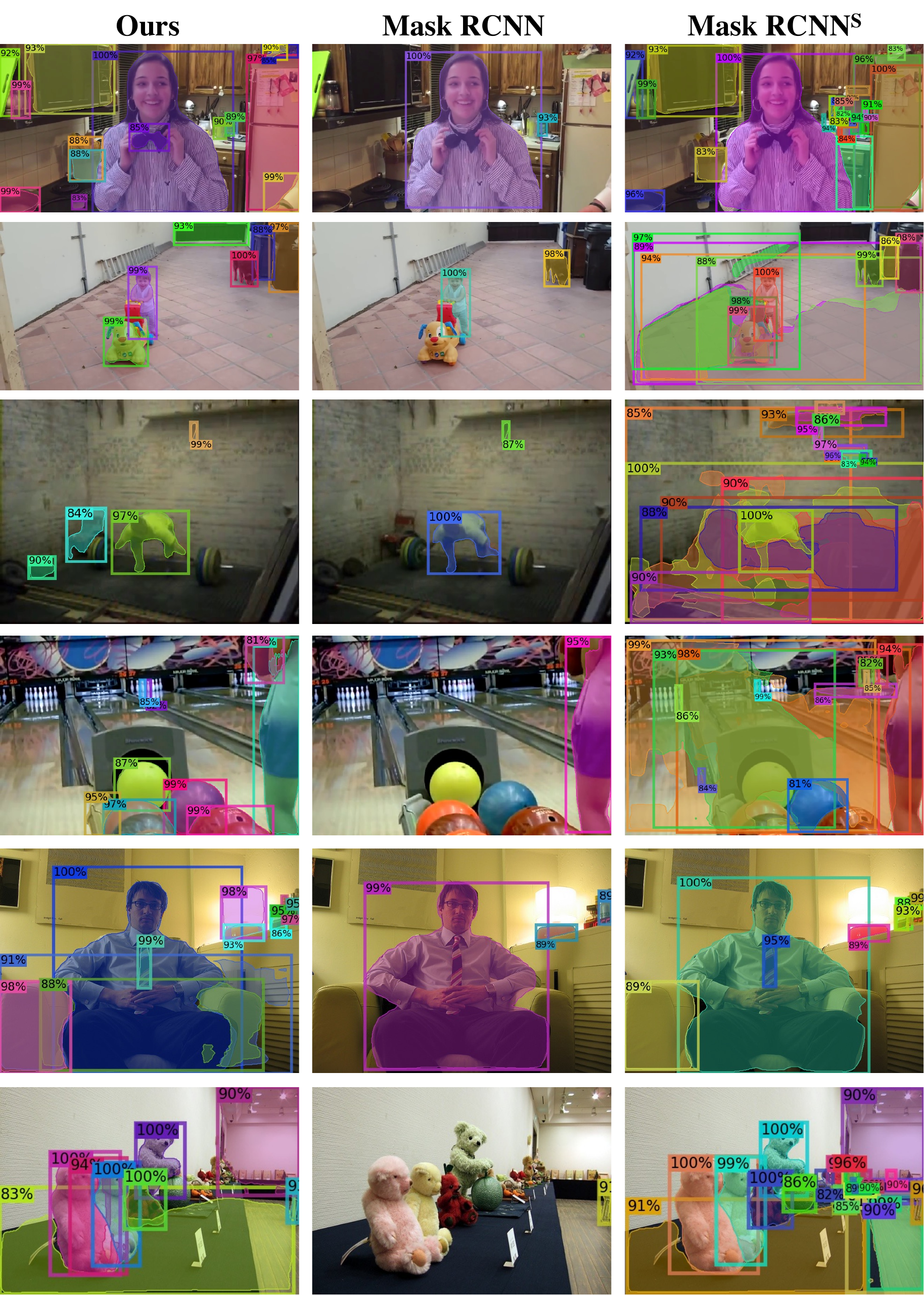}
    \vspace{-3mm}
    \caption{\small{\textbf{Visualization of models trained on COCO.} The images are from COCO and UVO.}}
    \label{fig:coco_all_to_uvo_appendix}
\end{figure*}

\begin{figure*}[t]
    \centering
    \href{https://cs-people.bu.edu/keisaito/videos/video_let/video1_concat.mp4}{\includegraphics[width=0.99\textwidth]{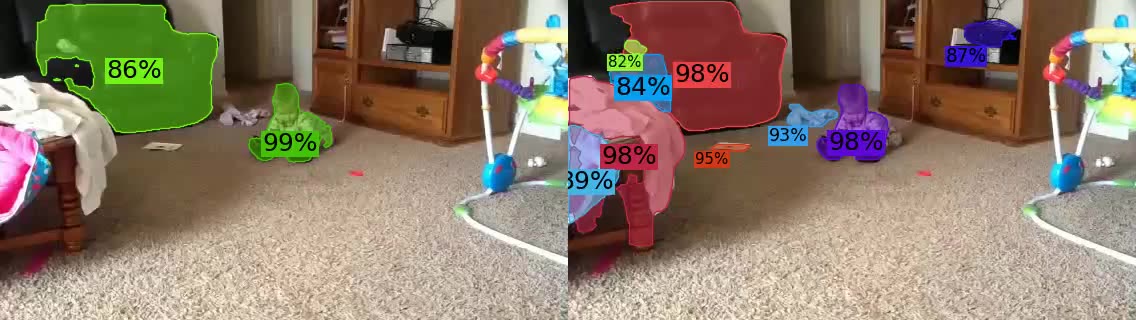}}
    %\embedvideo{\includegraphics[width=\linewidth]{images//video1_thumnail.jpg}}{videos/video1_concat.mp4}
    \vspace{-3mm}
%    \caption{\small{\textbf{Video demo of models trained on COCO. Left: Mask RCNN. Right: \ours. Use Adobe Acrobat to play the video.}}}
 \caption{\small{\textbf{Video demo of models trained on COCO. Left: Mask RCNN. Right: \ours. Click the image to play the video.}}}
    \label{fig:video1}
\end{figure*}

\begin{figure*}[t]
    \centering
   \href{https://cs-people.bu.edu/keisaito/videos/video_let/video2_concat.mp4}{\includegraphics[width=0.99\textwidth]{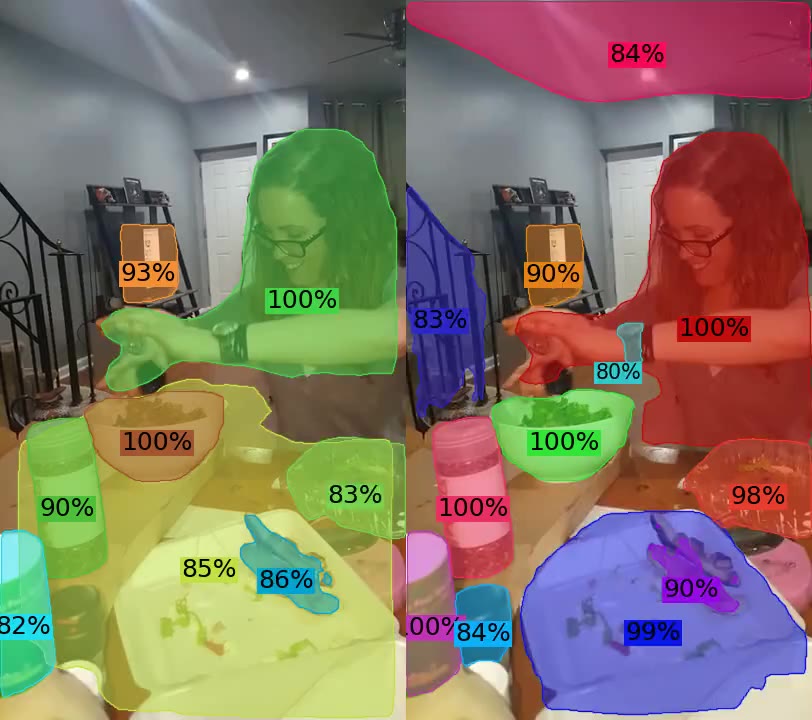}}

  %  \embedvideo{\includegraphics[width=0.7\linewidth]{images/video2_thumnail.jpg}}{videos/video2_concat.mp4}
% arxiv submission
   %\caption{\small{\textbf{Video demo of models trained on COCO. Left: Mask RCNN. Right: \ours.Click the image to play the video.}}}
%% supplemental submission
    \caption{\small{\textbf{Video demo of models trained on COCO. Left: Mask RCNN. Right: \ours.}}}
    \label{fig:video2}
\end{figure*}

\section{Visualization}

\noindent\textbf{Cityscapes.} 
Fig.~\ref{fig:cityscape_examples} visualizes some qualitative results. Leftmost two images are from the validation set of Cityscapes, others are from Mapillary. We see that, as indicated by the quantitative results, \ours detects more objects, \eg, \textit{baby carriage} in the leftmost image. However, it is also true that \ours misses novel objects such as \textit{dog} in the leftmost image, probably because there are no categories similar to dogs in the Cityscapes' 8 training categories. This fact indicates some room for improvement in our approach.

\noindent\textbf{More visualizations in COCO.} 
Fig.~\ref{fig:voc_to_coco_appendix} and \ref{fig:coco_all_to_uvo_appendix} are additional visualizations in VOC-COCO and COCO, respectively. Note that we add the results of $\text{Mask RCNN}^{\text{S}}$, which are not visualized in the main paper due to a limited space. $\text{Mask RCNN}^{\text{S}}$ locates many novel objects while generating many false positives. This is probably due to the imbalanced sampling of background regions. By contrast, \ours detects many novel objects, \eg, \textit{elephants, toilet paper, lizard, statue, toy}, \etc, with small number of false positives.

\noindent\textbf{Demo on video.} 
Fig.~\ref{fig:video1} and \ref{fig:video2} are demo of applying \ours to UVO~\cite{wang2021unidentified} videos.
Click the images to play the videos.
%These videos can be played with Adobe Acrobat or play videos attached with this appendix.

%\begin{center}
%        \embedvideo{\includegraphics[width=0.5\textwidth]{latex/images/video1_thumnail.jpg}}{latex/videos/video1_concat.mp4}
%\end{center}

\clearpage
\bibliographystyle{splncs04}
\bibliography{egbib}
\end{document}